\documentclass[sn-mathphys-ay]{sn-jnl}% Math and Physical Sciences Author Year Reference Style
\usepackage{graphicx}%
\usepackage{multirow}%
\usepackage{amsmath,amssymb,amsfonts}%
\usepackage{amsthm}%
\usepackage{mathrsfs}%
\usepackage[title]{appendix}%
\usepackage{xcolor}%
\usepackage{textcomp}%
\usepackage{manyfoot}%
\usepackage{booktabs}%
\usepackage{algorithm}%
\usepackage{algorithmicx}%
\usepackage{algpseudocode}%
\usepackage{listings}%
\theoremstyle{thmstyleone}%
\theoremstyle{thmstyletwo}%

\theoremstyle{thmstylethree}%

\begin{document}

\title[Article Title]{Cooperative SQL Generation for Segmented Databases By Using Multi-functional LLM Agents \footnote{This work was supported in part by the National Natural Science Foundation of China (Nos. 62272392 and U22A2025), the Key Research and Development Program of Shaanxi Province (No. 2023-YBGY-405), and Fundamental Research Funds for the Central Universities (No. G2023KY0603)}}

%%=============================================================%%
%% GivenName	-> \fnm{Joergen W.}
%% Particle	-> \spfx{van der} -> surname prefix
%% FamilyName	-> \sur{Ploeg}
%% Suffix	-> \sfx{IV}
%% \author*[1,2]{\fnm{Joergen W.} \spfx{van der} \sur{Ploeg} 
%%  \sfx{IV}}\email{iauthor@gmail.com}
%%=============================================================%%

\author[1,2]{\fnm{Zhiguang} \sur{Wu}}
%\email{zhiguangwu@nwpu.edu.cn}

%\email{shang@nwpu.edu.cn}
%\equalcont{These authors contributed equally to this work.}

\author[3]{\fnm{Fengbin} \sur{Zhu}}

\author[1,2]{\fnm{Xuequn} \sur{Shang}}

\author[1,2]{\fnm{Yupei} \sur{Zhang}}
\equalcont{\textit{Corresponding author: Yupei Zhang (ypzhaang@nwpu.edu.cn), Pan Zhou (panzhou@smu.edu.sg)}}

\author[4]{\fnm{Pan} \sur{Zhou}}

\equalcont{\textit{Corresponding author: Yupei Zhang (ypzhaang@nwpu.edu.cn), Pan Zhou (panzhou@smu.edu.sg)}}

\affil[1]{\orgdiv{School of Computer Science}, \orgname{Northwestern Polytechnical University}, \orgaddress{\street{Dongxiang Road}, \city{Xi'an}, \postcode{710129}, \state{Shaanxi}, \country{China}}}

\affil[2]{\orgdiv{Big Data Shortage and Management}, \orgname{MIIT}, \orgaddress{\street{Dongxiang Road}, \city{Xi'an}, \postcode{710129}, \state{Shaanxi}, \country{China}}}

\affil[3]{\orgdiv{School of Computing}, \orgname{National University of Singapore}, \orgaddress{\street{21 Lower Kent Ridge Road}, \city{Singapore}, \postcode{119077}, \state{Singapore}, \country{Singapore}}}

\affil[4]{\orgdiv{School of Computing and Information Systems}, \orgname{Singapore Management University }, \orgaddress{\street{80 Stamford Road}, \city{Singapore}, \postcode{178902}, \state{Singapore}, \country{Singapore}}}

%%==================================%%
%% Sample for unstructured abstract %%
%%==================================%%

\abstract{Text-to-SQL task aims to automatically yield SQL queries according to user text questions. To address this problem, we propose a \textbf{C}ooperative \textbf{S}QL Generation framework based on \textbf{M}ulti-functional \textbf{A}gents (CSMA) through information interaction among large language model (LLM) based agents who own part of the database schema seperately. Inspired by the collaboration in human teamwork, CSMA consists of  three stages: 1) Question-related schema collection, 2) Question-corresponding SQL query generation, and 3) SQL query correctness check. In the first stage, agents analyze their respective schema and communicate with each other to collect the schema information relevant to the question. In the second stage, agents try to generate the corresponding SQL query for the question using the collected information. In the third stage, agents check if the SQL query is created correctly according to their known information. This interaction-based method makes the question-relevant part of database schema from each agent to be used for SQL generation and check. Experiments on the Spider and Bird benckmark demonstrate that CSMA achieves a high performance level comparable to the state-of-the-arts, meanwhile holding the private data in these individual agents.}


%%\pacs[JEL Classification]{D8, H51}

%%\pacs[MSC Classification]{35A01, 65L10, 65L12, 65L20, 65L70}

\maketitle

\section{Introduction}\label{sec1}
Text-to-SQL refers to the task that creating the accurate corresponding SQL queries for natural language questions given by users, intended to make non-professional users easier to raise their questions through databases without knowledge on SQL, which has significant application value on a large variety of fields in our daily life, such as data intelligence, business analysis and other applications with database interaction. With the advent of LLM, researchers start to use them not only for many NLP tasks, but also to drive embodied agents to execute tasks collaborating with other agents or human in the 3D environment. Unfortunately, agents based on LLMs are rarely applied to Text-to-SQL task in spite of their remarkable capabilities of solving complicated problems in various areas \citep{zhang2020meta,zhang2022multi}.

\begin{figure}[h]
\centering
\includegraphics[width=0.9\textwidth]{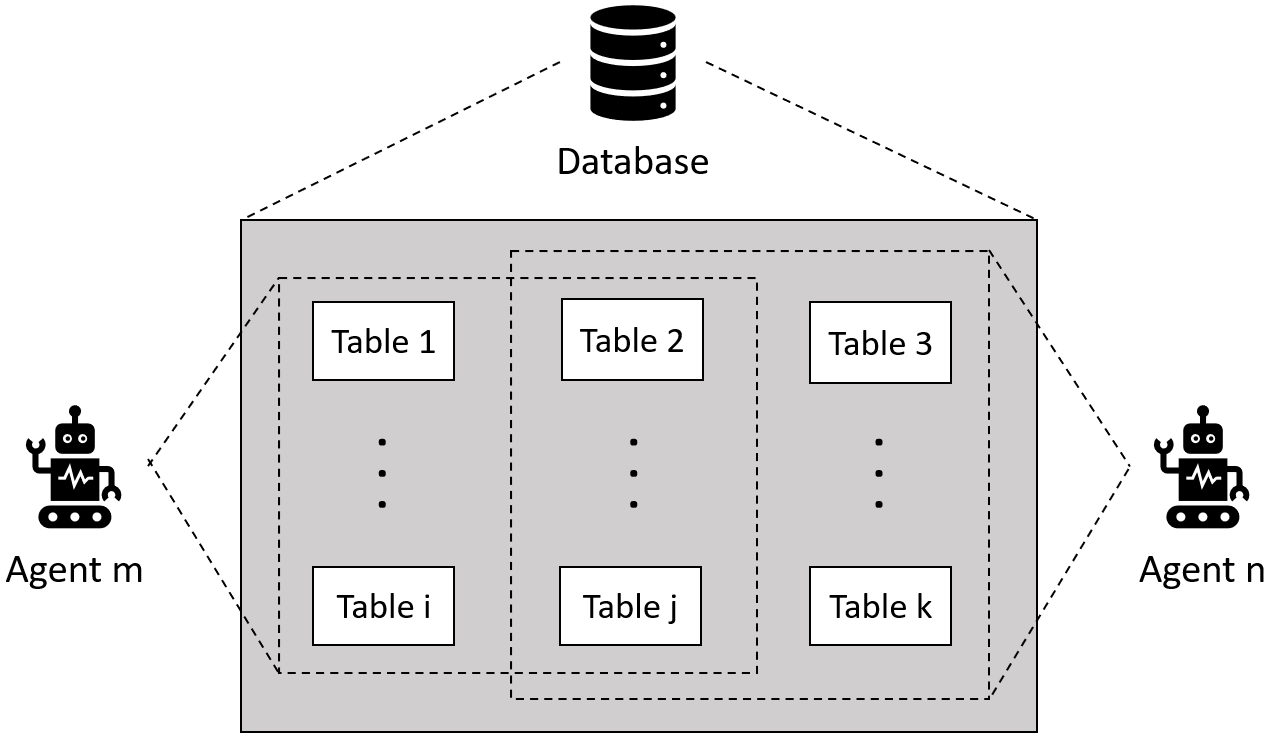}
\caption{The distribution of database tables for agents. Each agent masters a different part of the database which might overlap with each other.}\label{fig-1}
\end{figure}

Among the very scarce papers which combine LLM-based agents and Text-to-SQL, \citet{wang2024macsqlmultiagentcollaborativeframework} introduces a novel framework utilizing several LLM-based agents for this task, achieving a pretty high execution accuracy on the BIRD benchmark. These agents play a functional role in the whole sequential problem-solving process separately, sharing all the schema of the database, which is just suitable to the small-scale database when the problem of valid data proportion and data privacy is not need to be considered. However, in the real enterprise production environment, the large-scale relational databases usually cover information of many aspects, some of which nearly has nothing to do with others, which means creating SQL queries for user questions usually needs just a fraction of the whole database schema. Moreover, industrial database information is often allocated to different departments or employees according to their permission rather than distributed to only one individual, which is shown in Fig. \ref{fig-1}.

Considering the limitation of existing methods when faced with these realistic factors and illuminated by research on embodied agents, we propose a \textbf{C}ooerative \textbf{S}QL Generation framework based on \textbf{M}ulti-functional \textbf{A}gents (CSMA), containing multiple LLM-based multi-functional agents, who master only part of the database schema separately and need to interact with each other to obtain their probably missing information from others constantly and eventually create the accurate SQL queries for user questions in several stages.

In the first stage, agents need to maintain a global schema serving as an interaction bridge so as to break the gap caused by the initially different schema among them. In this way, each agent expands his schema knowledge with only schema related to the user-given question from others instead of the whole database schema, which increases the proportion of valid schema in their schema knowledge and make the irrelevant schema of an agent invisible to others. In addition, we design a retention mechanism for schema extraction, which limits the minimum column number of tables in the extraction result, and also clarify the specific rule of schema merging. Finally, the necessary schema agents acquire through the schema communication lays the solid foundation for the next stage.

In the second stage, due to the incompleteness and alteration of the schema knowledge of agents, the task for them is trying to generate SQL for the question from their own perspective iteratively. On the one hand, agents possess expertise in distinct segments of the database schema, enabling them to approach the generation task from entirely unique perspectives. On the other hand, the schema knowledge of them is continuously evolving due to the ongoing updates made by the working agent throughout each cycle. It is implemented by prompting LLMs with diversified strategies according to the question complexity and the parameter count of model. Lastly, a corresponding SQL is created by the selected working agent in each round, which provides the checking and judging materials for the next phase.

In the third stage, the SQL created in the second stage of each iteration is likely to be correct, we propose a technique for verifying the accuracy of the generated SQL and determining whether the procedure can be concluded. Since the SQL is crafted according to the known schema of agent in the current round, it is fed to the agent in the next round to check for the correctness. Then, we use the checking result to judge whether the whole process can be terminated, namely we stop the procedure if the result is positive.

In the experiment, we set up four conditions for comparison which are severally 1) one agent with partial schema, 2) one agent with whole schema, 3) two agents with partial schema and 4) two agents with whole schema. Evaluating them on the Spider and Bird benchmark shows that 3) nearly reach the same high performance as that of 2) and 4), which can prove the good effect of CSMA.

\section{Related Work}\label{sec2}
\subsection{Text-to-SQL}\label{subsec1}

Research in Text-to-SQL mainly goes through three stages. At first, it focus on using deep learning methods. A large proportion of them adopt sequence-to-sequence framework containing an encoder such as LSTM~\citep{yu2018typesqlknowledgebasedtypeawareneural}, Transformer~\citep{zhang2019editingbasedsqlquerygeneration, kelkar2020bertranddrimprovingtexttosqlusing} and GNN~\citep{yu2018syntaxsqlnetsyntaxtreenetworks, bogin2019representingschemastructuregraph} which transforms user questions from natural language to vectors and a decoder such as Grammer~\citep{cao2021lgesqllinegraphenhanced, hui2022s2sqlinjectingsyntaxquestionschema}, Sketch~\citep{yu2018typesqlknowledgebasedtypeawareneural, choi2020ryansqlrecursivelyapplyingsketchbased}, DSL~\citep{guo2019complextexttosqlcrossdomaindatabase, chen2021shadowgnngraphprojectionneural} and Re-Ranking~\citep{kelkar2020bertranddrimprovingtexttosqlusing} which is designed to create the corresponding SQL queries from user questions represented by vectors~\citep{qin2022surveytexttosqlparsingconcepts}. Later on, in the era of pretrained language models (PLMs) such as BERT~\citep{devlin2019bertpretrainingdeepbidirectional}, BART~\citep{lewis2019bartdenoisingsequencetosequencepretraining} and T5~\citep{raffel2023exploringlimitstransferlearning}, performances of this task improve a lot due to abundant knowledge extracted from massive corpus, though slightly affected by intrinsically different distribution between natural language questions and database descriptions which are founded by \citet{yin2020tabertpretrainingjointunderstanding} and \citet{yu2021grappagrammaraugmentedpretrainingtable}. Recently with the rise of large language models (LLMs), plenty of works have tried to apply them to this task, showing their impressive capacity and potential~\citep{liu2023comprehensiveevaluationchatgptszeroshot, pourreza2023dinsqldecomposedincontextlearning, wang2024dbcopilotscalingnaturallanguage}.

\subsection{Embodied Agent}\label{subsec2}
Considering outstanding abilities of LLMs in understanding natural language, generating dialog content and reasoning complex prolems, in some papers of recent years, LLMs are used to drive not only single embodied agent, but also multiple agents to perform a task in an environment that resembles human habitation collaboratively. For example, \citet{huang2022languagemodelszeroshotplanners} leverages the knowledge acquired by LLMs from interactive environments to generate more correct and executable plans for an embodied agent. Facing more complex tasks, \citet{song2023llmplannerfewshotgroundedplanning} empowers LLMs to instruct an embodied agent to finish them with perception of the surrounding environments from a long-term view. Inspired by the general collaboration in human teamwork, there are some studies harnessing the capabilities of LLMs to build multiple embodied agents who work with each other cooperatively towards a common goal increasing efficiency to a large extent, such as \citet{zhang2024buildingcooperativeembodiedagents}. Different from the relatively straightforward tasks in these works, \citet{simateam2024scalinginstructableagentssimulated} devote themselves to developing a comprehensive embodied system containing more intelligent agents who can follow the instructions in any form to accomplish challenging tasks in 3D environments.

\section{Preliminaries}\label{sec3}

We define the setting of CSMA as using decentralized partially observable Markov decision process (DEC-POMDP) \citep{Spaan2006DecentralizedPU} to complete the Text-to-SQL task \citep{wang2024macsqlmultiagentcollaborativeframework}. The relevant variables are defined as follows:
\begin{itemize}
\item The $n$ agents are denoted as $I = \{I_1, ..., I_n\}$.
\item The database is denoted as $D$.
\item The natural language question given by users is denoted as $Q$.
\item The optionally provided external knowledge which is helpful to this task is denoted as $K$.
\item The generated SQL query corresponding to question $Q$ is denoted as $Y$.
\item The schema of database $D$ is denoted as $S = \{T, C\}$, where $T = \{T_1, ..., T_{|T|}\}$ represents the tables in database $D$ and $C = \{C_1, ..., C_{|C|}\}$ represents the columns in tables $T$. $S_g$ represents the global schema shared among $n$ agents. $S_k = S_{k_1} \times ... \times S_{k_n} \ $represents the known schema of agents. $S_p = S_{p_1} \times ... \times S_{p_n} \ $represents the private schema owned by agents.
\item The whole action set is denoted as $A$. The action set of agent $I_i$ is denoted as $A_i = A_i^e \times A_i^s \times A_i^c$, where $A_i^e$ is the set of schema level actions, $A_i^s$ is the set of SQL generation actions and $A_i^c$ is the set of SQL checking actions.
\end{itemize}

The state transition of global schema depends on agents' private schema and their schema level actions. The probability for the global schema in the k iteration is as follow:
\begin{equation}
    \label{eqn-1}
    p(s_g|s_p, a) = \prod \limits_{i=1}^k p(s_g^i|s_g^{i-1}, s_{p_{t(i)}}, a_{t(i)}^e),
\end{equation}
where $s_g \in S_g$, $s_p \in S_p$, $a \in A$, $S_g^i$ represents the global schema after i iterations and function $t(\cdot)$ decides which agent to use for work in the current iteration.

The reward function to the team for the state transition of global schema is as follows:
\begin{equation}
    \label{eqn-2}
    R_{e}(s, a) = \sum \limits_{i=1}^k -C(a_{t(i)}^e) + J_e(s_g^{i}, q) - J_e(s_g^{i-1}, q)
\end{equation}
where $q \in Q$, $C(\cdot)$ calculates the cost of action $A$ and $J_e(\cdot)$ calculates the matching score between the global schema $S_g$ and the question $Q$.

The state transition of SQL query depends on the global schema, agents' private schema and their SQL generation actions. The probability for the SQL query in the k iteration is as follow:
\begin{equation}
    \label{eqn-3}
    p(y|s_g, s_p, a) = \prod \limits_{i=1}^k p(s_g^i|s_g^{i-1}, s_{p_{t(i)}}, a_{t(i)}^e) \cdot p(y^i|s_g^{i}, s_{p_{t(i)}}, a_{t(i)}^s) \cdot J_s(y^i, s_{k_{n(i)}}, q),
\end{equation}
where $n(\cdot)$ decides which agent to use in the next iteration and $J_s(\cdot)$ judges if the generated SQL $Y$ is acceptable according to agent's known schema $S_k$ and question $Q$.

The reward function to the team for the state transition of SQL query is as follows:
\begin{equation}
    \label{eqn-4}
    R_{s}(y, s, a) = \sum \limits_{i=1}^k -C^{'}(a, i) + J_s(y^{i+1}, s_{k_{n(i+1)}}, q) - J_s(y^{i}, s_{k_{n(i)}}, q)
\end{equation}
where $y \in Y$ and $C^{'}(a, i) = C(a_{t(i)}^e) + C(a_{t(i)}^s) + C(a_{n(i)}^c)$.

So the overall reward function to the team is:
\begin{equation}
    \label{eqn-5}
    R = \lambda_{e}R_{e}(s, a) + \lambda_{s}R_{s}(y, s, a)
\end{equation}
where $\lambda_{e}$ and $\lambda_{s}$ represents the weight of state transition of schema and SQL respectively.

\section{Cooperative SQL Generation by Multi-functional LLM Agents}\label{sec4}

We first introduce the overall framework of Cooperative SQL Generation by Multi-functional Agents (CSMA) in Sec. \ref{sec-4.1} and then elaborate on the three stages of CSMA, namely question-related schema collection, question-corresponding SQL generation and SQL correctness check, in Sec. \ref{sec-4.2}, Sec. \ref{sec-4.3} and Sec. \ref{sec-4.4} respectively.

\subsection{Overall Framework of CSMA}
\label{sec-4.1}
CSMA is intended to generate SQL queries for natural language questions given by users according to a segmented database using LLM-based multi-functional agents. The whole process usually requires multiple rounds, in each of which, the selected agent performs preset procedures until the generated SQL is judged as correct by the agent selected in the next one.

In CSMA, each agent holds a private schema which is a subset of database schema and has several functions, including schema extraction, schema merging, SQL generation and SQL checking. With schema extraction function, agents can extract tables, columns, primary and foreign key relationship relevant to the user-given question from schema. Schema merging means agents can merge multiple sub schema into an overall schema. SQL generation function can be used to generate the corresponding SQL query for the question according to specific schema by agents. SQL checking refers to the function that agents can check if the SQL is correct on the syntactic and semantic level given a question and schema.

\begin{figure}[h]
\begin{center}
\includegraphics[width=0.9\textwidth]{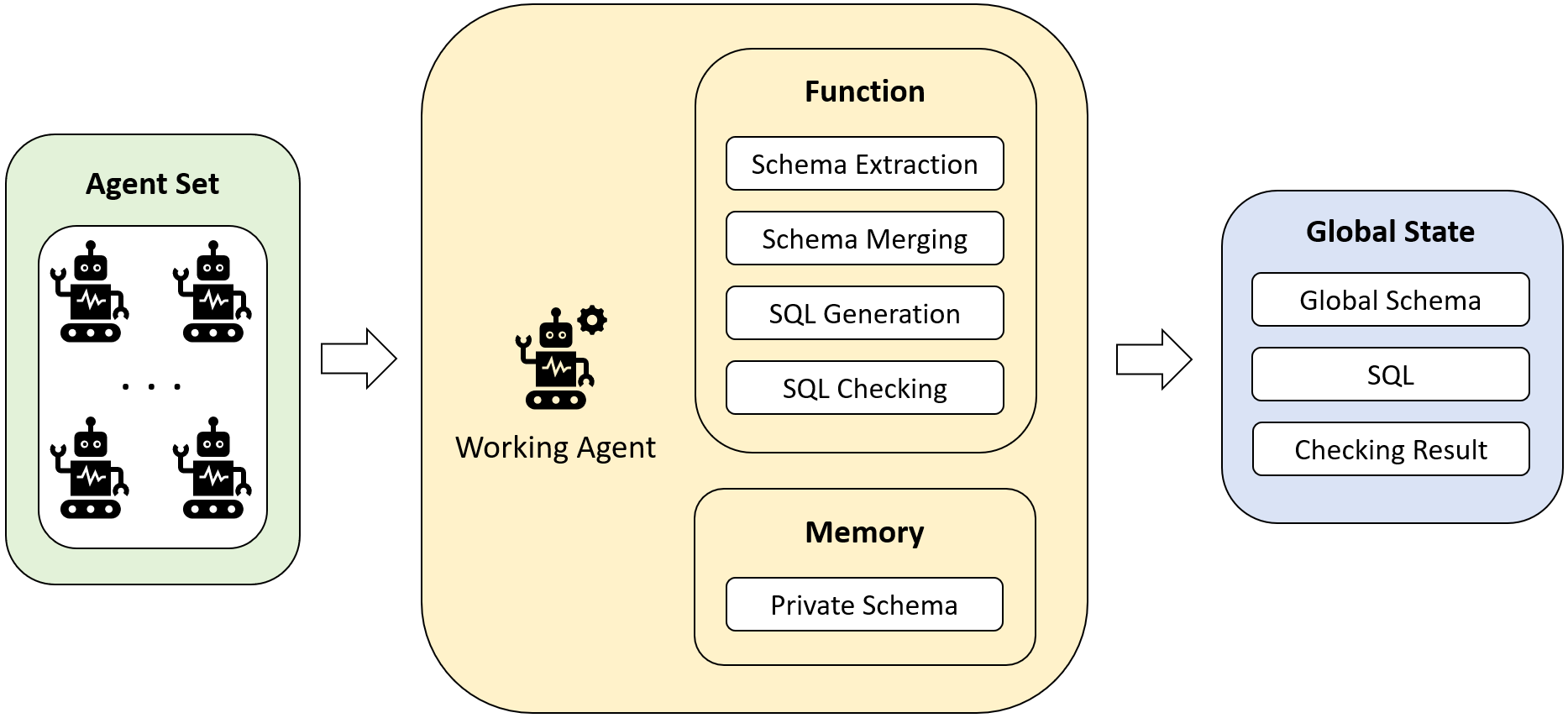}
\end{center}
\caption{The overview of our \textbf{CSMA} framework, where in each round an working agent is selected to interact with the global state.}
\label{fig-2}
\end{figure}

\begin{figure}[h]
\begin{center}
\includegraphics[width=0.9\textwidth]{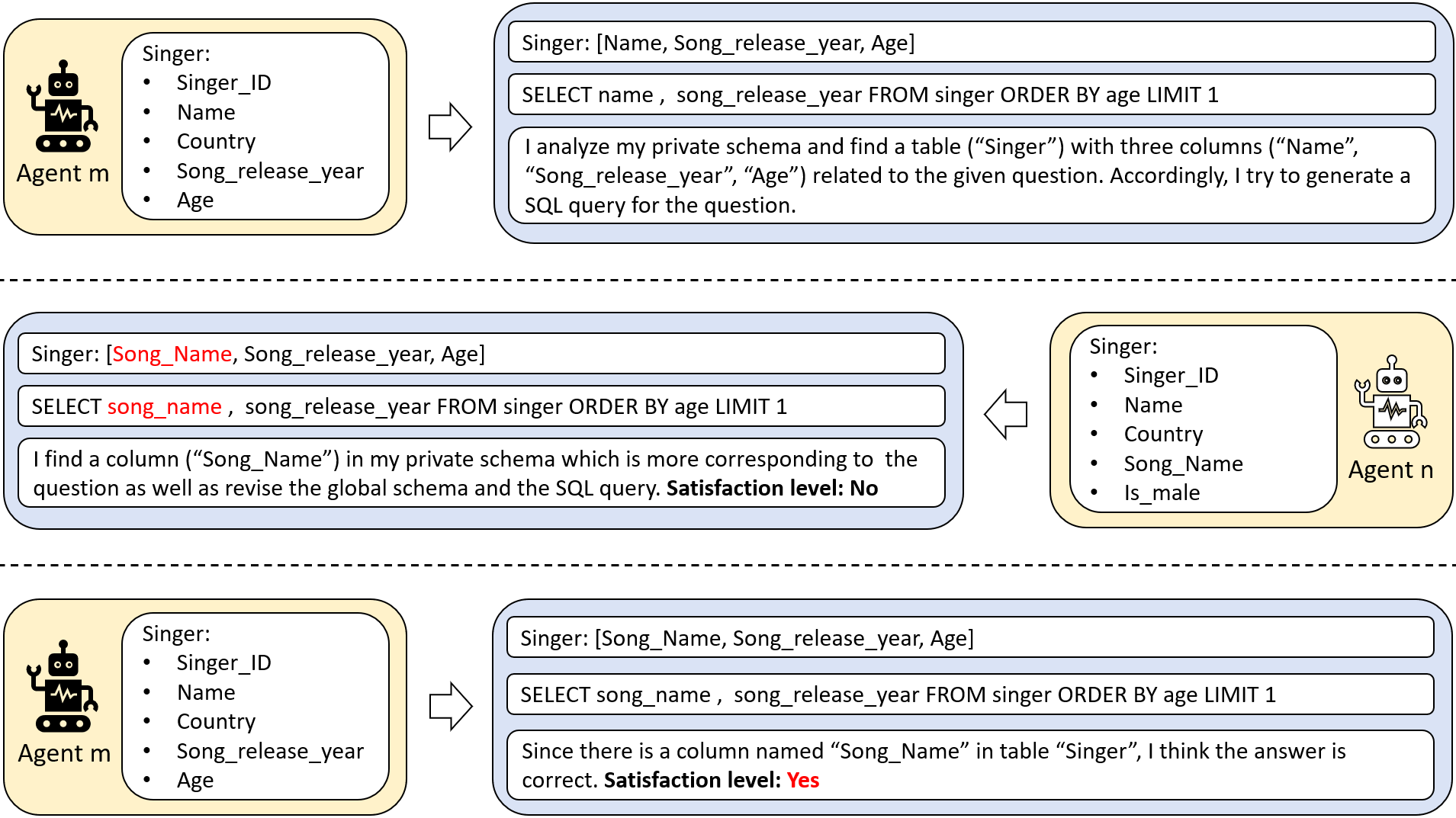}
\end{center}
\caption{An example of the whole Text-to-SQL process. Several agents take turns to update the global states until the checking result is positive.}
\label{fig-5}
\end{figure}

To ensure the effectiveness and efficiency of task completion, CSMA is designed to be a three-stage process shown in Fig. \ref{fig-2}. In the first stage, to collect the schema which is necessary for generating SQL, agents are required to analyze their private schema to maintain a global schema shared among them commonly. In the second stage, agents' known schema is extended from only their private schema to including the global schema. Combining their private schema and the global schema, they need to analyze and extract the part which is useful to generating SQL for the given question. In the third stage, to judge the correctness of generated SQL queries, they are fed to and checked carefully by agents according to agents' known schema at that moment. Then the process can be terminated if the checking result is positive, otherwise enter into the next cycle. Fig. \ref{fig-5} gives a concrete example about the entire process. In the following subsections, we will take turns to elaborate on these stages.

\subsection{Question-related Schema Collection}
\label{sec-4.2}

Due to that the database is segmented into several parts, which are held by each agent separately, we need to have all the relevant schema at first before generating SQL for the question. Therefore, in this stage, we design a global schema which is shared and maintained by all the agents.

\begin{figure}[h]
\begin{center}
\includegraphics[width=0.9\textwidth]{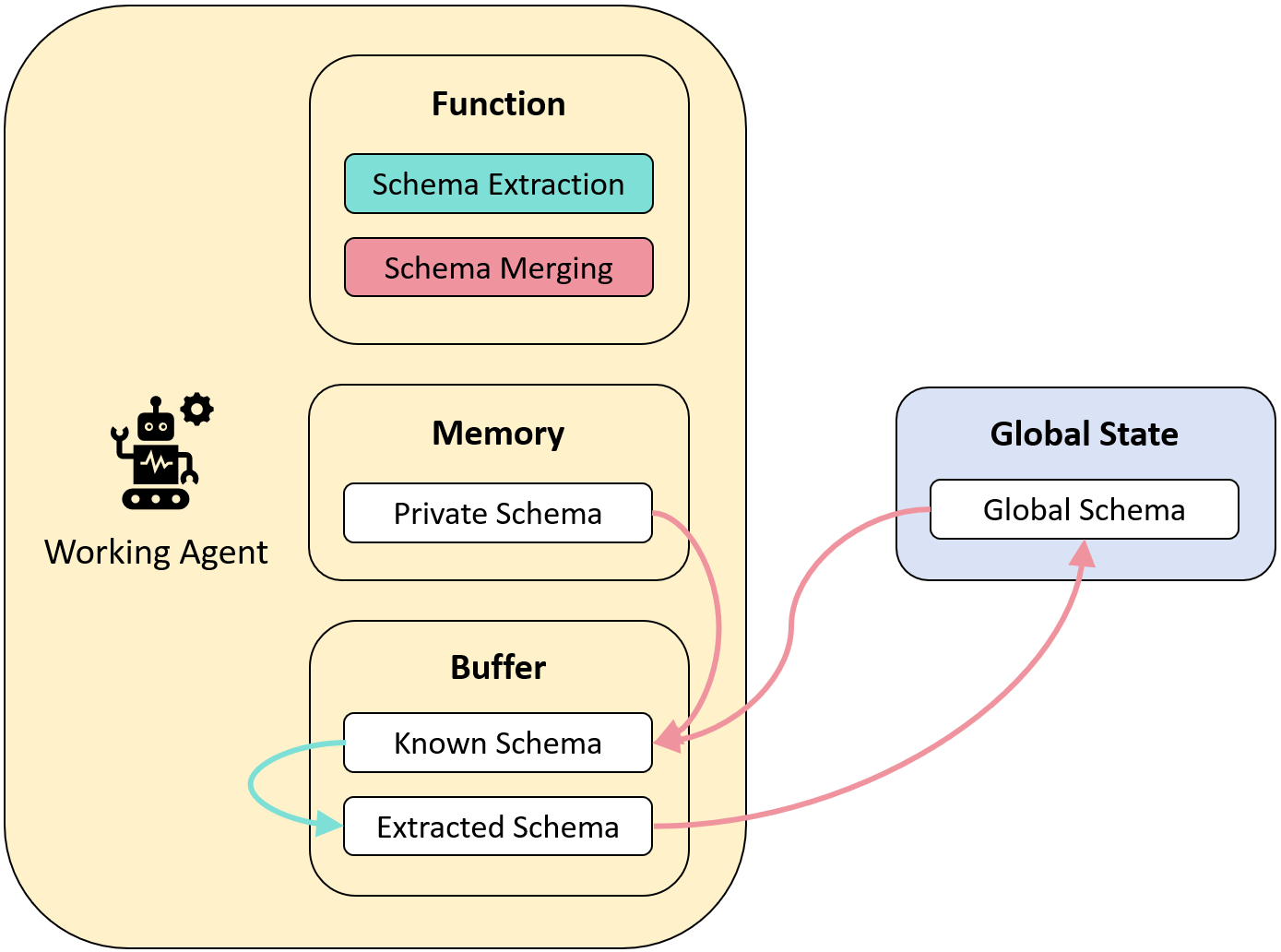}
\end{center}
\caption{The procedure of question-related schema collection. The private schema and global schema are merged into the known schema, which is extracted and merged into the global schema.}
\label{fig-3}
\end{figure}

The global schema is initialized to be empty and updated dynamically by each agent in turn until the whole process is terminated. As shown in Fig. \ref{fig-3}, to be specific, in each iteration, the agent working in this round uses the schema merging function to merge his private schema and the global schema into an intermediate schema, which is named as his known schema at this moment. Next, the necessary part of his known schema is extracted by his schema extraction function after comprehensively analyzing and merged with the global schema to make it updated. In this way, the global schema updated constantly serves as a medium for agents to exchange their respective schema, which makes the most relevant part to be collected for subsequent stages while making sure the unnecessary part does not be focused.

There are also two key points which need to be clarified additionally. The first one is the retention mechanism of schema extraction. In order to make up for the schema loss caused by schema extraction, a threshold $\delta$ is introduced to decide the minimum count of columns retained for each table at least, which means $\delta$ columns of a table should be retained even if the retention count is judged to be fewer than $\delta$. Secondly, for two schema, the merging rule is that if a table only exists in one of them, just keep it and its columns as well as if a table exists in both of them, keep it and the union of its columns. These points are both designed to make agents utilize more comprehensive database schema which is found significant in our experiments.

\subsection{Question-corresponding SQL Generation}
\label{sec-4.3}
The global schema, maintained by all the agents in multiple rounds, provides them with additional schema from others which does not belong to them initially. But the update of it in each round makes the known schema of agents change constantly. So in this stage we propose an iterative method to make agents generate SQL according to their latest known schema.

\begin{figure}[h]
\begin{center}
\includegraphics[width=0.9\textwidth]{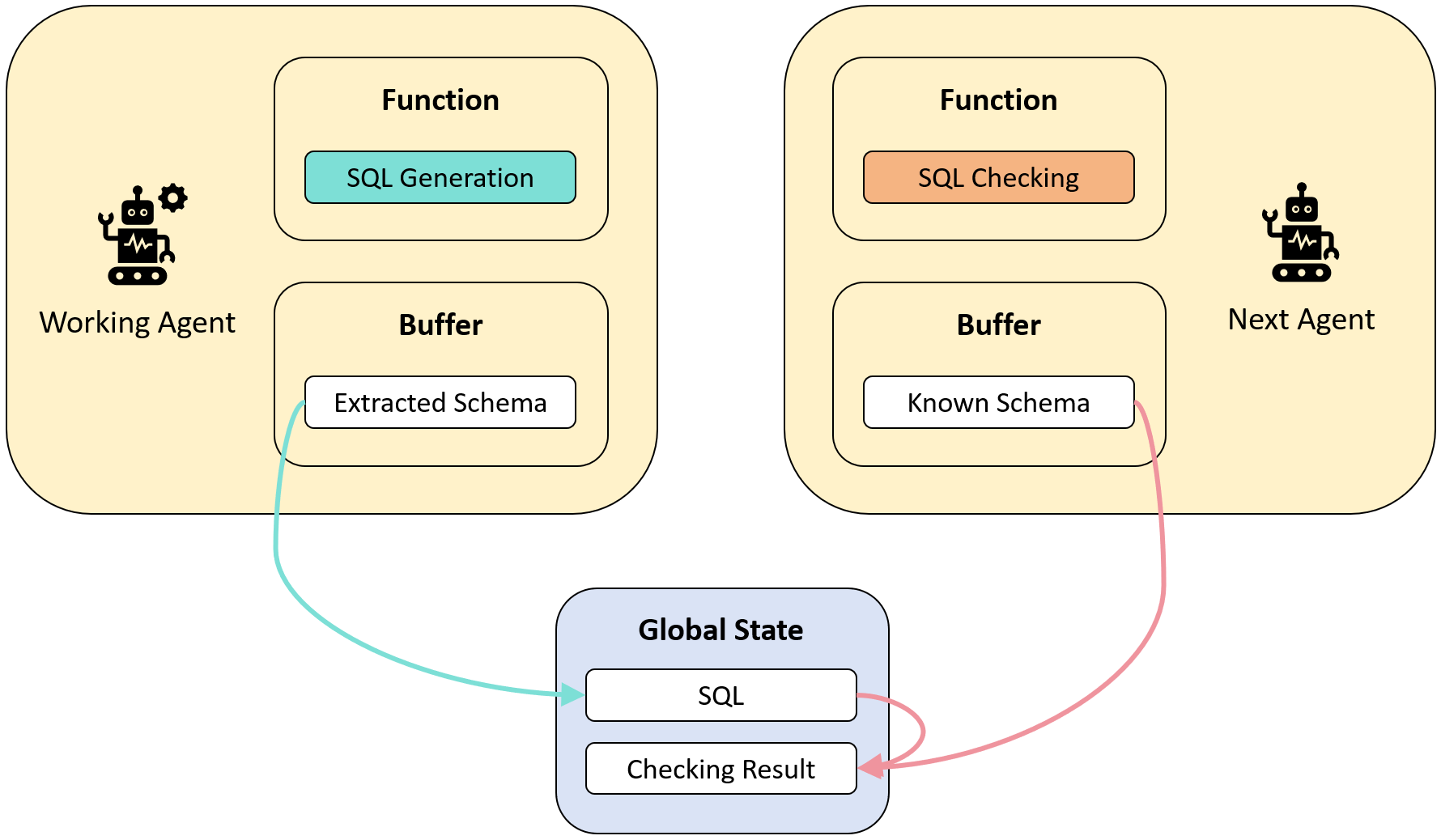}
\end{center}
\caption{The procedure of question-corresponding SQL generation and SQL correctness check. The working agent uses the SQL Generation function to generate the SQL query, which is checked by the next agent according to its known schema.}
\label{fig-4}
\end{figure}

As illustrated in Fig. \ref{fig-4}, specifically, in each round, after a series of maintaining operations to global schema in the last stage, the working agent get the extracted part from his known schema by the way. Then he needs to try generating SQL for question using the SQL generation function based on his extracted schema. However, there might be missing information in his extracted schema which is necessary, it probably takes several attempts for the generation process. In an overall view, each agent masters different part of the database schema, which makes them perform the generation action from a totally different point of view, although the global schema expands the overlap among the known schema of agents objectively. From the perspective of single agent, his known schema also keeps changing constantly due to the update of global schema by the working agent in each iteration. Therefore, the question-corresponding SQL needs to be created alternately by the agents with different known schema. On the other hand, it is also the process of choosing the most optimal schema state.

As for the implementation, there are multiple strategies which can be used to prompt LLMs to comprehend the question and analyze the schema accordingly. For example, we can design prompt to reason based on the question and schema directly or decompose the question into several sub-questions which are solved progressively following \citet{wang2024macsqlmultiagentcollaborativeframework}. Choosing between them mainly depends on the difficulty of question and the capability of LLM. When evaluating the difficulty of question, we treat the structure complexity of its golden SQL as criteria, including the count of tables involved in the "JOIN" clause, queries nested in the SQL and so on. In the condition of using LLMs with more parameters like gpt-3.5-turbo-16k and gpt-4, we tend to adopt the first prompt strategy to solve complex questions and the second strategy when facing easy questions. While for the case of using fewer-parameter LLMs such as gpt-3.5-turbo-0125, it is better to employ the second prompt strategy, whose performance even exceeds that of the first one. The reason is possibly that fewer parameters limit their ability to exert the powerful effect of the first one.

\subsection{SQL Correctness Check}
\label{sec-4.4}
Question-corresponding SQL is generated iteratively by the working agent of each round in the last stage, but its correctness is uncertain. Moreover, another significant issue is to decide when to stop in the process of iteration. So in this stage we introduce a method to check the correctness of generated SQL and judge if the process can be terminated.

The SQL is generated based on the known schema of currently working agent. Therefore, we allocate the correctness checking task to the agent who works in the next round and adopt the checking result as the criterion for whether to end the iteration, which is shown in Fig. \ref{fig-4}. Specifically, after performing schema-related actions in the first stage, the working agent needs to use the SQL checking function to check if the SQL generated in the last round is correct according to its latest known schema and carefully give his judging result. At last, the whole process can be stopped if he gives a positive result, otherwise it continues and the currently working agent goes into the SQL generation stage.

\section{Experiments}
\subsection{Setup}
\subsubsection{Datasets}
The Spider dataset, published by \citet{yu-etal-2018-spider}, including 8,034 question-query pairs regarding 200 unique databases and 138 different domains, is often used to measure the capabilities of text-to-SQL parsers adjusting to new and previously unseen database schema. 

The BIRD dataset, introduced by \citet{li2023llmservedatabaseinterface}, sets a more robust benchmark, containing more large-scale databases and external knowledge, which brings new challenges for text-to-SQL translation.

In this research, we evaluate the performance of our framework (CSMA) using the development and test set of Spider and the test set of BIRD. 

\subsubsection{Evaluation Metrics}
Following \citet{zhong2020semanticevaluationtexttosqldistilled}, we use two metrics, execution accuracy (EX) and valid efficiency score (VES) to evaluate the effect of CSMA compared to the baseline described in Sec. \ref{sec-5.1.3} in this work. Execution Accuracy (EX) is the measure of the percentage of questions within the assessment set where the outcomes of the executed predicted queries match exactly with those of the actual queries, compared to the total quantity of questions assessed. The Valid Efficiency Score (VES) is formulated to evaluate the effectiveness of the valid SQL statements produced by the models. It is necessary to clarify that "valid SQLs" denote the forecasted SQL queries that yield result sets equivalent to those of the corresponding ground-truth SQLs.

\subsubsection{Baseline}
\label{sec-5.1.3}

Existing methods usually adopt various strategies to improve the performance of Text-to-SQL, but all of them complete the task on the condition of reasoning according to the whole database schema directly. For example, in \citet{wang2024macsqlmultiagentcollaborativeframework}, the selector directly extract the relevant schema of a large database, which is used by the decomposer to generate SQL for the complex question and the refiner to refine the final SQL with mistakes. In the experiments of this work, we select it as the baseline to evaluate the effect of CSMA.

\subsection{Main Performance}
To evaluate the performance of our setting, we set up four conditions for the Text-to-SQL task, named as "OnePart", "OneAll", "TwoPart" and "TwoAll". "OnePart" represents there is one agent who only masters part of the schema. "OneAll" represents there is one agent who holds the whole schema, which is corresponding to the setting of plenty of existing methods. "TwoPart" represents there are two agents who have part of the schema, overlapping with each other possibly, corresponding to the setting of our method. “TwoAll” represents there are two agents both mastering the whole schema. Regarding the distributing manner, we randomly select half of tables in the schema and allocate them and all of their columns to the agent for "OnePart", while for "TwoPart" we divide the entire schema into two pieces with the same number of tables which are held by two agents severally.

\begin{table}[h]
\caption{Execution accuracy(EX) and Valid efficiency score(VES) of four settings on both dev and test set of Spider and the dev set of BIRD.}
\label{tab-1}
\begin{tabular}{lllllll}
\toprule
 \multirow{3}{*}{\textbf{Condition}} & \multicolumn{4}{c}{\textbf{Spider}} & \multicolumn{2}{c}{\textbf{BIRD}} \\ 
\cmidrule(lr){2-5} \cmidrule(lr){6-7}
& \multicolumn{2}{c}{\textbf{Dev}} & \multicolumn{2}{c}{\textbf{Test}} & \multicolumn{2}{c}{\textbf{Dev}} \\
\cmidrule(lr){2-3} \cmidrule(lr){4-5} \cmidrule(r){6-7}
& \textbf{EM} & \textbf{EX} & \textbf{EM} & \textbf{EX} & \textbf{EX} & \textbf{VES} \\
\midrule
OnePart & 19.34 & 42.55 & 23.61 & 45.65 & 25.36 & 11.09 \\
OneAll & 31.62 & 75.15 & 32.18 & 74.57 & 48.89 & 20.62 \\
TwoPart & 32.40 & \textbf{75.92} & 33.44 & \textbf{74.94} & \textbf{49.28} & 20.74 \\
TwoAll & 30.95 & 76.31 & 32.70 & 74.90 & 49.35 & 20.76 \\
\bottomrule
\end{tabular}
\end{table}

We conduct experiments on the dev and test set of Spider as well as the dev set of BIRD on four conditions respectively. The measure result of EM and EX on Spider as well as EX and VES on BIRD is shown in Table \ref{tab-1}. Obviously, the performance of "OneAll", "TwoPart" and "TwoAll" exceeds that of "OnePart" drastically on both benchmarks because the incomplete schema makes the agent can not generate the accurate SQL. It is also reasonable that "OneAll" and "TwoAll" achieve the best result compared to "OnePart" and "TwoPart" since the complete schema is surely beneficial to SQL creation. Furthermore, we can find that the effect of "TwoAll"  outperforms that of "OneAll" with the assistance of another agent in most cases except for just a little lower on the test set of Spider. The most significant and notable is that the result of "TwoPart" can reach almost the same level as that of "OneAll" and "TwoAll", which demonstrates in CSMA the agents with different schema can complement mutually by cooperating with each other, that is to say, CSMA largely achieves its intended objective.

\begin{table}[h]
\caption{Execution accuracy of different number of agents on the dev set of Spider and BIRD}
\label{tab-2}
\begin{tabular}{llllllllll}
\toprule
\multirow{2}{*}{\textbf{Method}} & \multicolumn{5}{c}{\textbf{Spider}} & \multicolumn{4}{c}{\textbf{BIRD}} \\ 
\cmidrule(lr){2-6} \cmidrule(lr){7-10}
& \textbf{Easy} & \textbf{Medium} & \textbf{Hard} & \textbf{Extra} & \textbf{All} & \textbf{Simple} & \textbf{Mod.} & \textbf{Chall.} & \textbf{All} \\
\midrule
Agent-2 & 96.15 & 78.21 & 77.14 & 70.00 & \textbf{78.77} & 68.31 & 56.94 & 42.62 & \textbf{53.89} \\
Agent-3 & 96.15 & 79.49 & 77.14 & 65.00 & 78.21 & 68.31 & 53.80 & 37.45 & 51.66 \\
Agent-4 & 92.31 & 80.77 & 77.14 & 60.00 & 77.09 & 66.69 & 51.22 & 33.72 & 48.57 \\
\bottomrule
\end{tabular}
\end{table}

Apart from experiments on four conditions above, we also do contrast experiment on two benchmarks BIRD and Spider, according to the number of agents, where we explore the effect of the number of agents on the performance of our framework. The experiment results are shown in Table \ref{tab-2}, which indicates that with the increase of the number of agents, the accuracy rate generally declined, although it fluctuated in different difficulties. In addition, compared with the accuracy of Spider, the decline trend of BIRD is more obvious, which may be caused by the relatively high difficult level of BIRD, which shows that the number of agents allocated is more important in the more difficult benchmark.

\subsection{Ablation Study}
Table \ref{tab-3} shows the results of an ablation study for the \textbf{CSMA} framework on the BIRD dev set. The table lists different variations of the \textbf{CSMA} framework, including with and without some components or mechanisms such as retention, exchange and checking. The other columns represent the accuracy of the framework on different levels of difficulty: Easy, Medium, Hard and Extra, as well as the overall accuracy.

\begin{table*}[h]
\caption{Execution accuracy of \textbf{CSMA} ablation study on the dev set of Spider and BIRD.}
\label{tab-3}
\resizebox{\textwidth}{!}{
\begin{tabular}{llllllllll}
\toprule
\multirow{2}{*}{\textbf{Method}} & \multicolumn{5}{c}{\textbf{Spider}} & \multicolumn{4}{c}{\textbf{BIRD}} \\ 
\cmidrule(lr){2-6} \cmidrule(lr){7-10}
& \textbf{Easy} & \textbf{Medium} & \textbf{Hard} & \textbf{Extra} & \textbf{All} & \textbf{Simple} & \textbf{Mod.} & \textbf{Chall.} & \textbf{All}\\
\midrule
CSMA & 96.15 & 78.21 & 77.14 & 70.00 & \textbf{78.77} & 68.52 & 57.17 & 42.82 & \textbf{54.05} \\
w/o retention & 96.15 & 82.05 & 65.71 & 52.50 & 74.30 & 68.11 & 59.96 & 34.37 & 50.56 \\
w/o exchange & 96.15 & 79.49 & 57.14 & 65.00 & 74.30 & 68.11 & 58.75 & 31.34 & 50.24 \\
w/o checking & 96.15 & 78.21 & 74.29 & 62.50 & 76.54 & 68.52 & 56.74 & 32.49 & 52.73 \\
\bottomrule
\end{tabular}
}
\end{table*}

The study results show that the original \textbf{CSMA} framework achieves an accuracy of 96.15\% on Easy, 78.21\% on Medium, 77.14\% on Hard, and 70.00\% on Extra, with an overall accuracy of 78.77\%. When removing the retention or exchange mechanism, the overall accuracy decreases to 74.30\%, in spite of a slight increase for Medium. Similarly, removing the checking component also leads to drop in the overall accuracy.

Overall, the ablation study shows that each component or mechanism of the \textbf{CSMA} framework plays a significant role in achieving high accuracy, as their removal results in decreased performance in the overall accuracy.

\subsection{Study on the Capability of In-context Learning}
Leveraging the in-context learning capability of LLMs, we designed an experiment to investigate the relationship between the number of examples in the prompt and the performance of our framework on two benchmarks. The results of this experiment are presented in Table \ref{tab-4}.

\begin{table}[h]
\caption{Evaluation of different number of shots on the dev set of Spider and BIRD}
\label{tab-4}
\begin{tabular}{lllll}
\toprule
\multirow{2}{*}{\textbf{Few-shot}} & \multicolumn{2}{c}{\textbf{Spider}} & \multicolumn{2}{c}{\textbf{BIRD}} \\ 
\cmidrule(lr){2-3} \cmidrule(lr){4-5}
& \textbf{EM} & \textbf{EX} & \textbf{EX} & \textbf{VES} \\
\midrule
0-shot & 31.82 & 75.64 & 48.84 & 20.81 \\
1-shot & 33.40 & 76.77 & 51.98 & 22.50 \\
2-shot & \textbf{34.27} & \textbf{80.14} & \textbf{55.92} & \textbf{23.37} \\
\bottomrule
\end{tabular}
\end{table}

The experimental results demonstrate that as the number of examples in the prompt increases from 0 to 2, the performance of our framework consistently improves across both benchmarks, BIRD and Spider. This indicates that our framework effectively leverages the examples provided in the prompt, enhancing its generalization capabilities with additional examples. Notably, the highest performance is achieved with 2-shot evaluation, suggesting that the model can learn efficiently from a limited number of examples. Therefore, this experiment validates that our framework, built upon LLMs, exhibits excellent in-context learning capabilities.

\section{Conclusion}
In summary, this paper introduces the \textbf{CSMA} framework, which adopts multi-functional agents to address challenges collaboratively in Text-to-SQL tasks for segmented databases. The framework achieves an execution accuracy of 74.94 when evaluated on the Spider benchmark, surpassing that of the other settings. This work presents a novel approach to Text-to-SQL achieving high execution accuracy and data efficiency at the same time in this domain, which can be used in industrial large-scale databases. In the future, we will develop this work for knowledge management in education \cite{zhang2023predicting} and extend into a federate learning schema for privacy protection \cite{zhang2023federated}.

\bibliography{sn-bibliography}% common bib file

%% BioMed_Central_Bib_Style_v1.01

\begin{thebibliography}{32}
% BibTex style file: bmc-mathphys.bst (version 2.1), 2014-07-24
\ifx \bisbn   \undefined \def \bisbn  #1{ISBN #1}\fi
\ifx \binits  \undefined \def \binits#1{#1}\fi
\ifx \bauthor  \undefined \def \bauthor#1{#1}\fi
\ifx \batitle  \undefined \def \batitle#1{#1}\fi
\ifx \bjtitle  \undefined \def \bjtitle#1{#1}\fi
\ifx \bvolume  \undefined \def \bvolume#1{\textbf{#1}}\fi
\ifx \byear  \undefined \def \byear#1{#1}\fi
\ifx \bissue  \undefined \def \bissue#1{#1}\fi
\ifx \bfpage  \undefined \def \bfpage#1{#1}\fi
\ifx \blpage  \undefined \def \blpage #1{#1}\fi
\ifx \burl  \undefined \def \burl#1{\textsf{#1}}\fi
\ifx \doiurl  \undefined \def \doiurl#1{\url{https://doi.org/#1}}\fi
\ifx \betal  \undefined \def \betal{\textit{et al.}}\fi
\ifx \binstitute  \undefined \def \binstitute#1{#1}\fi
\ifx \binstitutionaled  \undefined \def \binstitutionaled#1{#1}\fi
\ifx \bctitle  \undefined \def \bctitle#1{#1}\fi
\ifx \beditor  \undefined \def \beditor#1{#1}\fi
\ifx \bpublisher  \undefined \def \bpublisher#1{#1}\fi
\ifx \bbtitle  \undefined \def \bbtitle#1{#1}\fi
\ifx \bedition  \undefined \def \bedition#1{#1}\fi
\ifx \bseriesno  \undefined \def \bseriesno#1{#1}\fi
\ifx \blocation  \undefined \def \blocation#1{#1}\fi
\ifx \bsertitle  \undefined \def \bsertitle#1{#1}\fi
\ifx \bsnm \undefined \def \bsnm#1{#1}\fi
\ifx \bsuffix \undefined \def \bsuffix#1{#1}\fi
\ifx \bparticle \undefined \def \bparticle#1{#1}\fi
\ifx \barticle \undefined \def \barticle#1{#1}\fi
\bibcommenthead
\ifx \bconfdate \undefined \def \bconfdate #1{#1}\fi
\ifx \botherref \undefined \def \botherref #1{#1}\fi
\ifx \url \undefined \def \url#1{\textsf{#1}}\fi
\ifx \bchapter \undefined \def \bchapter#1{#1}\fi
\ifx \bbook \undefined \def \bbook#1{#1}\fi
\ifx \bcomment \undefined \def \bcomment#1{#1}\fi
\ifx \oauthor \undefined \def \oauthor#1{#1}\fi
\ifx \citeauthoryear \undefined \def \citeauthoryear#1{#1}\fi
\ifx \endbibitem  \undefined \def \endbibitem {}\fi
\ifx \bconflocation  \undefined \def \bconflocation#1{#1}\fi
\ifx \arxivurl  \undefined \def \arxivurl#1{\textsf{#1}}\fi
\csname PreBibitemsHook\endcsname

%%% 1
\bibitem[\protect\citeauthoryear{Bogin et~al.}{2019}]{bogin2019representingschemastructuregraph}
\begin{botherref}
\oauthor{\bsnm{Bogin}, \binits{B.}},
\oauthor{\bsnm{Gardner}, \binits{M.}},
\oauthor{\bsnm{Berant}, \binits{J.}}:
Representing Schema Structure with Graph Neural Networks for Text-to-SQL Parsing
(2019).
\url{https://arxiv.org/abs/1905.06241}
\end{botherref}
\endbibitem

%%% 2
\bibitem[\protect\citeauthoryear{Cao et~al.}{2021}]{cao2021lgesqllinegraphenhanced}
\begin{botherref}
\oauthor{\bsnm{Cao}, \binits{R.}},
\oauthor{\bsnm{Chen}, \binits{L.}},
\oauthor{\bsnm{Chen}, \binits{Z.}},
\oauthor{\bsnm{Zhao}, \binits{Y.}},
\oauthor{\bsnm{Zhu}, \binits{S.}},
\oauthor{\bsnm{Yu}, \binits{K.}}:
LGESQL: Line Graph Enhanced Text-to-SQL Model with Mixed Local and Non-Local Relations
(2021).
\url{https://arxiv.org/abs/2106.01093}
\end{botherref}
\endbibitem

%%% 3
\bibitem[\protect\citeauthoryear{Chen et~al.}{2021}]{chen2021shadowgnngraphprojectionneural}
\begin{botherref}
\oauthor{\bsnm{Chen}, \binits{Z.}},
\oauthor{\bsnm{Chen}, \binits{L.}},
\oauthor{\bsnm{Zhao}, \binits{Y.}},
\oauthor{\bsnm{Cao}, \binits{R.}},
\oauthor{\bsnm{Xu}, \binits{Z.}},
\oauthor{\bsnm{Zhu}, \binits{S.}},
\oauthor{\bsnm{Yu}, \binits{K.}}:
ShadowGNN: Graph Projection Neural Network for Text-to-SQL Parser
(2021).
\url{https://arxiv.org/abs/2104.04689}
\end{botherref}
\endbibitem

%%% 4
\bibitem[\protect\citeauthoryear{Choi et~al.}{2020}]{choi2020ryansqlrecursivelyapplyingsketchbased}
\begin{botherref}
\oauthor{\bsnm{Choi}, \binits{D.}},
\oauthor{\bsnm{Shin}, \binits{M.C.}},
\oauthor{\bsnm{Kim}, \binits{E.}},
\oauthor{\bsnm{Shin}, \binits{D.R.}}:
RYANSQL: Recursively Applying Sketch-based Slot Fillings for Complex Text-to-SQL in Cross-Domain Databases
(2020).
\url{https://arxiv.org/abs/2004.03125}
\end{botherref}
\endbibitem

%%% 5
\bibitem[\protect\citeauthoryear{Devlin et~al.}{2019}]{devlin2019bertpretrainingdeepbidirectional}
\begin{botherref}
\oauthor{\bsnm{Devlin}, \binits{J.}},
\oauthor{\bsnm{Chang}, \binits{M.-W.}},
\oauthor{\bsnm{Lee}, \binits{K.}},
\oauthor{\bsnm{Toutanova}, \binits{K.}}:
BERT: Pre-training of Deep Bidirectional Transformers for Language Understanding
(2019).
\url{https://arxiv.org/abs/1810.04805}
\end{botherref}
\endbibitem

%%% 6
\bibitem[\protect\citeauthoryear{Guo et~al.}{2019}]{guo2019complextexttosqlcrossdomaindatabase}
\begin{botherref}
\oauthor{\bsnm{Guo}, \binits{J.}},
\oauthor{\bsnm{Zhan}, \binits{Z.}},
\oauthor{\bsnm{Gao}, \binits{Y.}},
\oauthor{\bsnm{Xiao}, \binits{Y.}},
\oauthor{\bsnm{Lou}, \binits{J.-G.}},
\oauthor{\bsnm{Liu}, \binits{T.}},
\oauthor{\bsnm{Zhang}, \binits{D.}}:
Towards Complex Text-to-SQL in Cross-Domain Database with Intermediate Representation
(2019).
\url{https://arxiv.org/abs/1905.08205}
\end{botherref}
\endbibitem

%%% 7
\bibitem[\protect\citeauthoryear{Huang et~al.}{2022}]{huang2022languagemodelszeroshotplanners}
\begin{botherref}
\oauthor{\bsnm{Huang}, \binits{W.}},
\oauthor{\bsnm{Abbeel}, \binits{P.}},
\oauthor{\bsnm{Pathak}, \binits{D.}},
\oauthor{\bsnm{Mordatch}, \binits{I.}}:
Language Models as Zero-Shot Planners: Extracting Actionable Knowledge for Embodied Agents
(2022).
\url{https://arxiv.org/abs/2201.07207}
\end{botherref}
\endbibitem

%%% 8
\bibitem[\protect\citeauthoryear{Hui et~al.}{2022}]{hui2022s2sqlinjectingsyntaxquestionschema}
\begin{botherref}
\oauthor{\bsnm{Hui}, \binits{B.}},
\oauthor{\bsnm{Geng}, \binits{R.}},
\oauthor{\bsnm{Wang}, \binits{L.}},
\oauthor{\bsnm{Qin}, \binits{B.}},
\oauthor{\bsnm{Li}, \binits{B.}},
\oauthor{\bsnm{Sun}, \binits{J.}},
\oauthor{\bsnm{Li}, \binits{Y.}}:
S$^2$SQL: Injecting Syntax to Question-Schema Interaction Graph Encoder for Text-to-SQL Parsers
(2022).
\url{https://arxiv.org/abs/2203.06958}
\end{botherref}
\endbibitem

%%% 9
\bibitem[\protect\citeauthoryear{Kelkar et~al.}{2020}]{kelkar2020bertranddrimprovingtexttosqlusing}
\begin{botherref}
\oauthor{\bsnm{Kelkar}, \binits{A.}},
\oauthor{\bsnm{Relan}, \binits{R.}},
\oauthor{\bsnm{Bhardwaj}, \binits{V.}},
\oauthor{\bsnm{Vaichal}, \binits{S.}},
\oauthor{\bsnm{Khatri}, \binits{C.}},
\oauthor{\bsnm{Relan}, \binits{P.}}:
Bertrand-DR: Improving Text-to-SQL using a Discriminative Re-ranker
(2020).
\url{https://arxiv.org/abs/2002.00557}
\end{botherref}
\endbibitem

%%% 10
\bibitem[\protect\citeauthoryear{Li et~al.}{2023}]{li2023llmservedatabaseinterface}
\begin{botherref}
\oauthor{\bsnm{Li}, \binits{J.}},
\oauthor{\bsnm{Hui}, \binits{B.}},
\oauthor{\bsnm{Qu}, \binits{G.}},
\oauthor{\bsnm{Yang}, \binits{J.}},
\oauthor{\bsnm{Li}, \binits{B.}},
\oauthor{\bsnm{Li}, \binits{B.}},
\oauthor{\bsnm{Wang}, \binits{B.}},
\oauthor{\bsnm{Qin}, \binits{B.}},
\oauthor{\bsnm{Cao}, \binits{R.}},
\oauthor{\bsnm{Geng}, \binits{R.}},
\oauthor{\bsnm{Huo}, \binits{N.}},
\oauthor{\bsnm{Zhou}, \binits{X.}},
\oauthor{\bsnm{Ma}, \binits{C.}},
\oauthor{\bsnm{Li}, \binits{G.}},
\oauthor{\bsnm{Chang}, \binits{K.C.C.}},
\oauthor{\bsnm{Huang}, \binits{F.}},
\oauthor{\bsnm{Cheng}, \binits{R.}},
\oauthor{\bsnm{Li}, \binits{Y.}}:
Can LLM Already Serve as A Database Interface? A BIg Bench for Large-Scale Database Grounded Text-to-SQLs
(2023).
\url{https://arxiv.org/abs/2305.03111}
\end{botherref}
\endbibitem

%%% 11
\bibitem[\protect\citeauthoryear{Liu et~al.}{2023}]{liu2023comprehensiveevaluationchatgptszeroshot}
\begin{botherref}
\oauthor{\bsnm{Liu}, \binits{A.}},
\oauthor{\bsnm{Hu}, \binits{X.}},
\oauthor{\bsnm{Wen}, \binits{L.}},
\oauthor{\bsnm{Yu}, \binits{P.S.}}:
A comprehensive evaluation of ChatGPT's zero-shot Text-to-SQL capability
(2023).
\url{https://arxiv.org/abs/2303.13547}
\end{botherref}
\endbibitem

%%% 12
\bibitem[\protect\citeauthoryear{Lewis et~al.}{2019}]{lewis2019bartdenoisingsequencetosequencepretraining}
\begin{botherref}
\oauthor{\bsnm{Lewis}, \binits{M.}},
\oauthor{\bsnm{Liu}, \binits{Y.}},
\oauthor{\bsnm{Goyal}, \binits{N.}},
\oauthor{\bsnm{Ghazvininejad}, \binits{M.}},
\oauthor{\bsnm{Mohamed}, \binits{A.}},
\oauthor{\bsnm{Levy}, \binits{O.}},
\oauthor{\bsnm{Stoyanov}, \binits{V.}},
\oauthor{\bsnm{Zettlemoyer}, \binits{L.}}:
BART: Denoising Sequence-to-Sequence Pre-training for Natural Language Generation, Translation, and Comprehension
(2019).
\url{https://arxiv.org/abs/1910.13461}
\end{botherref}
\endbibitem

%%% 13
\bibitem[\protect\citeauthoryear{Pourreza and Rafiei}{2023}]{pourreza2023dinsqldecomposedincontextlearning}
\begin{botherref}
\oauthor{\bsnm{Pourreza}, \binits{M.}},
\oauthor{\bsnm{Rafiei}, \binits{D.}}:
DIN-SQL: Decomposed In-Context Learning of Text-to-SQL with Self-Correction
(2023).
\url{https://arxiv.org/abs/2304.11015}
\end{botherref}
\endbibitem

%%% 14
\bibitem[\protect\citeauthoryear{Qin et~al.}{2022}]{qin2022surveytexttosqlparsingconcepts}
\begin{botherref}
\oauthor{\bsnm{Qin}, \binits{B.}},
\oauthor{\bsnm{Hui}, \binits{B.}},
\oauthor{\bsnm{Wang}, \binits{L.}},
\oauthor{\bsnm{Yang}, \binits{M.}},
\oauthor{\bsnm{Li}, \binits{J.}},
\oauthor{\bsnm{Li}, \binits{B.}},
\oauthor{\bsnm{Geng}, \binits{R.}},
\oauthor{\bsnm{Cao}, \binits{R.}},
\oauthor{\bsnm{Sun}, \binits{J.}},
\oauthor{\bsnm{Si}, \binits{L.}},
\oauthor{\bsnm{Huang}, \binits{F.}},
\oauthor{\bsnm{Li}, \binits{Y.}}:
A Survey on Text-to-SQL Parsing: Concepts, Methods, and Future Directions
(2022).
\url{https://arxiv.org/abs/2208.13629}
\end{botherref}
\endbibitem

%%% 15
\bibitem[\protect\citeauthoryear{Raffel et~al.}{2023}]{raffel2023exploringlimitstransferlearning}
\begin{botherref}
\oauthor{\bsnm{Raffel}, \binits{C.}},
\oauthor{\bsnm{Shazeer}, \binits{N.}},
\oauthor{\bsnm{Roberts}, \binits{A.}},
\oauthor{\bsnm{Lee}, \binits{K.}},
\oauthor{\bsnm{Narang}, \binits{S.}},
\oauthor{\bsnm{Matena}, \binits{M.}},
\oauthor{\bsnm{Zhou}, \binits{Y.}},
\oauthor{\bsnm{Li}, \binits{W.}},
\oauthor{\bsnm{Liu}, \binits{P.J.}}:
Exploring the Limits of Transfer Learning with a Unified Text-to-Text Transformer
(2023).
\url{https://arxiv.org/abs/1910.10683}
\end{botherref}
\endbibitem

%%% 16
\bibitem[\protect\citeauthoryear{Spaan et~al.}{2006}]{Spaan2006DecentralizedPU}
\begin{bchapter}
\bauthor{\bsnm{Spaan}, \binits{M.T.J.}},
\bauthor{\bsnm{Gordon}, \binits{G.J.}},
\bauthor{\bsnm{Vlassis}, \binits{N.A.}}:
\bctitle{Decentralized planning under uncertainty for teams of communicating agents}.
In: \bbtitle{Adaptive Agents and Multi-Agent Systems}
(\byear{2006}).
\burl{https://api.semanticscholar.org/CorpusID:1751957}
\end{bchapter}
\endbibitem

%%% 17
\bibitem[\protect\citeauthoryear{Song et~al.}{2023}]{song2023llmplannerfewshotgroundedplanning}
\begin{botherref}
\oauthor{\bsnm{Song}, \binits{C.H.}},
\oauthor{\bsnm{Wu}, \binits{J.}},
\oauthor{\bsnm{Washington}, \binits{C.}},
\oauthor{\bsnm{Sadler}, \binits{B.M.}},
\oauthor{\bsnm{Chao}, \binits{W.-L.}},
\oauthor{\bsnm{Su}, \binits{Y.}}:
LLM-Planner: Few-Shot Grounded Planning for Embodied Agents with Large Language Models
(2023).
\url{https://arxiv.org/abs/2212.04088}
\end{botherref}
\endbibitem

%%% 18
\bibitem[\protect\citeauthoryear{Team et~al.}{2024}]{simateam2024scalinginstructableagentssimulated}
\begin{botherref}
\oauthor{\bsnm{Team}, \binits{S.}},
\oauthor{\bsnm{Raad}, \binits{M.A.}},
\oauthor{\bsnm{Ahuja}, \binits{A.}},
\oauthor{\bsnm{Barros}, \binits{C.}},
\oauthor{\bsnm{Besse}, \binits{F.}},
\oauthor{\bsnm{Bolt}, \binits{A.}},
\oauthor{\bsnm{Bolton}, \binits{A.}},
\oauthor{\bsnm{Brownfield}, \binits{B.}},
\oauthor{\bsnm{Buttimore}, \binits{G.}},
\oauthor{\bsnm{Cant}, \binits{M.}},
\oauthor{\bsnm{Chakera}, \binits{S.}},
\oauthor{\bsnm{Chan}, \binits{S.C.Y.}},
\oauthor{\bsnm{Clune}, \binits{J.}},
\oauthor{\bsnm{Collister}, \binits{A.}},
\oauthor{\bsnm{Copeman}, \binits{V.}},
\oauthor{\bsnm{Cullum}, \binits{A.}},
\oauthor{\bsnm{Dasgupta}, \binits{I.}},
\oauthor{\bsnm{Cesare}, \binits{D.}},
\oauthor{\bsnm{Trapani}, \binits{J.D.}},
\oauthor{\bsnm{Donchev}, \binits{Y.}},
\oauthor{\bsnm{Dunleavy}, \binits{E.}},
\oauthor{\bsnm{Engelcke}, \binits{M.}},
\oauthor{\bsnm{Faulkner}, \binits{R.}},
\oauthor{\bsnm{Garcia}, \binits{F.}},
\oauthor{\bsnm{Gbadamosi}, \binits{C.}},
\oauthor{\bsnm{Gong}, \binits{Z.}},
\oauthor{\bsnm{Gonzales}, \binits{L.}},
\oauthor{\bsnm{Gupta}, \binits{K.}},
\oauthor{\bsnm{Gregor}, \binits{K.}},
\oauthor{\bsnm{Hallingstad}, \binits{A.O.}},
\oauthor{\bsnm{Harley}, \binits{T.}},
\oauthor{\bsnm{Haves}, \binits{S.}},
\oauthor{\bsnm{Hill}, \binits{F.}},
\oauthor{\bsnm{Hirst}, \binits{E.}},
\oauthor{\bsnm{Hudson}, \binits{D.A.}},
\oauthor{\bsnm{Hudson}, \binits{J.}},
\oauthor{\bsnm{Hughes-Fitt}, \binits{S.}},
\oauthor{\bsnm{Rezende}, \binits{D.J.}},
\oauthor{\bsnm{Jasarevic}, \binits{M.}},
\oauthor{\bsnm{Kampis}, \binits{L.}},
\oauthor{\bsnm{Ke}, \binits{R.}},
\oauthor{\bsnm{Keck}, \binits{T.}},
\oauthor{\bsnm{Kim}, \binits{J.}},
\oauthor{\bsnm{Knagg}, \binits{O.}},
\oauthor{\bsnm{Kopparapu}, \binits{K.}},
\oauthor{\bsnm{Lampinen}, \binits{A.}},
\oauthor{\bsnm{Legg}, \binits{S.}},
\oauthor{\bsnm{Lerchner}, \binits{A.}},
\oauthor{\bsnm{Limont}, \binits{M.}},
\oauthor{\bsnm{Liu}, \binits{Y.}},
\oauthor{\bsnm{Loks-Thompson}, \binits{M.}},
\oauthor{\bsnm{Marino}, \binits{J.}},
\oauthor{\bsnm{Cussons}, \binits{K.M.}},
\oauthor{\bsnm{Matthey}, \binits{L.}},
\oauthor{\bsnm{Mcloughlin}, \binits{S.}},
\oauthor{\bsnm{Mendolicchio}, \binits{P.}},
\oauthor{\bsnm{Merzic}, \binits{H.}},
\oauthor{\bsnm{Mitenkova}, \binits{A.}},
\oauthor{\bsnm{Moufarek}, \binits{A.}},
\oauthor{\bsnm{Oliveira}, \binits{V.}},
\oauthor{\bsnm{Oliveira}, \binits{Y.}},
\oauthor{\bsnm{Openshaw}, \binits{H.}},
\oauthor{\bsnm{Pan}, \binits{R.}},
\oauthor{\bsnm{Pappu}, \binits{A.}},
\oauthor{\bsnm{Platonov}, \binits{A.}},
\oauthor{\bsnm{Purkiss}, \binits{O.}},
\oauthor{\bsnm{Reichert}, \binits{D.}},
\oauthor{\bsnm{Reid}, \binits{J.}},
\oauthor{\bsnm{Richemond}, \binits{P.H.}},
\oauthor{\bsnm{Roberts}, \binits{T.}},
\oauthor{\bsnm{Ruscoe}, \binits{G.}},
\oauthor{\bsnm{Elias}, \binits{J.S.}},
\oauthor{\bsnm{Sandars}, \binits{T.}},
\oauthor{\bsnm{Sawyer}, \binits{D.P.}},
\oauthor{\bsnm{Scholtes}, \binits{T.}},
\oauthor{\bsnm{Simmons}, \binits{G.}},
\oauthor{\bsnm{Slater}, \binits{D.}},
\oauthor{\bsnm{Soyer}, \binits{H.}},
\oauthor{\bsnm{Strathmann}, \binits{H.}},
\oauthor{\bsnm{Stys}, \binits{P.}},
\oauthor{\bsnm{Tam}, \binits{A.C.}},
\oauthor{\bsnm{Teplyashin}, \binits{D.}},
\oauthor{\bsnm{Terzi}, \binits{T.}},
\oauthor{\bsnm{Vercelli}, \binits{D.}},
\oauthor{\bsnm{Vujatovic}, \binits{B.}},
\oauthor{\bsnm{Wainwright}, \binits{M.}},
\oauthor{\bsnm{Wang}, \binits{J.X.}},
\oauthor{\bsnm{Wang}, \binits{Z.}},
\oauthor{\bsnm{Wierstra}, \binits{D.}},
\oauthor{\bsnm{Williams}, \binits{D.}},
\oauthor{\bsnm{Wong}, \binits{N.}},
\oauthor{\bsnm{York}, \binits{S.}},
\oauthor{\bsnm{Young}, \binits{N.}}:
Scaling Instructable Agents Across Many Simulated Worlds
(2024).
\url{https://arxiv.org/abs/2404.10179}
\end{botherref}
\endbibitem

%%% 19
\bibitem[\protect\citeauthoryear{Wang et~al.}{2024}]{wang2024dbcopilotscalingnaturallanguage}
\begin{botherref}
\oauthor{\bsnm{Wang}, \binits{T.}},
\oauthor{\bsnm{Lin}, \binits{H.}},
\oauthor{\bsnm{Han}, \binits{X.}},
\oauthor{\bsnm{Sun}, \binits{L.}},
\oauthor{\bsnm{Chen}, \binits{X.}},
\oauthor{\bsnm{Wang}, \binits{H.}},
\oauthor{\bsnm{Zeng}, \binits{Z.}}:
DBCopilot: Scaling Natural Language Querying to Massive Databases
(2024).
\url{https://arxiv.org/abs/2312.03463}
\end{botherref}
\endbibitem

%%% 20
\bibitem[\protect\citeauthoryear{Wang et~al.}{2024}]{wang2024macsqlmultiagentcollaborativeframework}
\begin{botherref}
\oauthor{\bsnm{Wang}, \binits{B.}},
\oauthor{\bsnm{Ren}, \binits{C.}},
\oauthor{\bsnm{Yang}, \binits{J.}},
\oauthor{\bsnm{Liang}, \binits{X.}},
\oauthor{\bsnm{Bai}, \binits{J.}},
\oauthor{\bsnm{Chai}, \binits{L.}},
\oauthor{\bsnm{Yan}, \binits{Z.}},
\oauthor{\bsnm{Zhang}, \binits{Q.-W.}},
\oauthor{\bsnm{Yin}, \binits{D.}},
\oauthor{\bsnm{Sun}, \binits{X.}},
\oauthor{\bsnm{Li}, \binits{Z.}}:
MAC-SQL: A Multi-Agent Collaborative Framework for Text-to-SQL
(2024).
\url{https://arxiv.org/abs/2312.11242}
\end{botherref}
\endbibitem

%%% 21
\bibitem[\protect\citeauthoryear{Yu et~al.}{2018}]{yu2018typesqlknowledgebasedtypeawareneural}
\begin{botherref}
\oauthor{\bsnm{Yu}, \binits{T.}},
\oauthor{\bsnm{Li}, \binits{Z.}},
\oauthor{\bsnm{Zhang}, \binits{Z.}},
\oauthor{\bsnm{Zhang}, \binits{R.}},
\oauthor{\bsnm{Radev}, \binits{D.}}:
TypeSQL: Knowledge-based Type-Aware Neural Text-to-SQL Generation
(2018).
\url{https://arxiv.org/abs/1804.09769}
\end{botherref}
\endbibitem

%%% 22
\bibitem[\protect\citeauthoryear{Yin et~al.}{2020}]{yin2020tabertpretrainingjointunderstanding}
\begin{botherref}
\oauthor{\bsnm{Yin}, \binits{P.}},
\oauthor{\bsnm{Neubig}, \binits{G.}},
\oauthor{\bsnm{Yih}, \binits{W.-t.}},
\oauthor{\bsnm{Riedel}, \binits{S.}}:
TaBERT: Pretraining for Joint Understanding of Textual and Tabular Data
(2020).
\url{https://arxiv.org/abs/2005.08314}
\end{botherref}
\endbibitem

%%% 23
\bibitem[\protect\citeauthoryear{Yu et~al.}{2021}]{yu2021grappagrammaraugmentedpretrainingtable}
\begin{botherref}
\oauthor{\bsnm{Yu}, \binits{T.}},
\oauthor{\bsnm{Wu}, \binits{C.-S.}},
\oauthor{\bsnm{Lin}, \binits{X.V.}},
\oauthor{\bsnm{Wang}, \binits{B.}},
\oauthor{\bsnm{Tan}, \binits{Y.C.}},
\oauthor{\bsnm{Yang}, \binits{X.}},
\oauthor{\bsnm{Radev}, \binits{D.}},
\oauthor{\bsnm{Socher}, \binits{R.}},
\oauthor{\bsnm{Xiong}, \binits{C.}}:
GraPPa: Grammar-Augmented Pre-Training for Table Semantic Parsing
(2021).
\url{https://arxiv.org/abs/2009.13845}
\end{botherref}
\endbibitem

%%% 24
\bibitem[\protect\citeauthoryear{Yu et~al.}{2018}]{yu2018syntaxsqlnetsyntaxtreenetworks}
\begin{botherref}
\oauthor{\bsnm{Yu}, \binits{T.}},
\oauthor{\bsnm{Yasunaga}, \binits{M.}},
\oauthor{\bsnm{Yang}, \binits{K.}},
\oauthor{\bsnm{Zhang}, \binits{R.}},
\oauthor{\bsnm{Wang}, \binits{D.}},
\oauthor{\bsnm{Li}, \binits{Z.}},
\oauthor{\bsnm{Radev}, \binits{D.}}:
SyntaxSQLNet: Syntax Tree Networks for Complex and Cross-DomainText-to-SQL Task
(2018).
\url{https://arxiv.org/abs/1810.05237}
\end{botherref}
\endbibitem

%%% 25
\bibitem[\protect\citeauthoryear{Yu et~al.}{2018}]{yu-etal-2018-spider}
\begin{bchapter}
\bauthor{\bsnm{Yu}, \binits{T.}},
\bauthor{\bsnm{Zhang}, \binits{R.}},
\bauthor{\bsnm{Yang}, \binits{K.}},
\bauthor{\bsnm{Yasunaga}, \binits{M.}},
\bauthor{\bsnm{Wang}, \binits{D.}},
\bauthor{\bsnm{Li}, \binits{Z.}},
\bauthor{\bsnm{Ma}, \binits{J.}},
\bauthor{\bsnm{Li}, \binits{I.}},
\bauthor{\bsnm{Yao}, \binits{Q.}},
\bauthor{\bsnm{Roman}, \binits{S.}},
\bauthor{\bsnm{Zhang}, \binits{Z.}},
\bauthor{\bsnm{Radev}, \binits{D.}}:
\bctitle{{S}pider: A large-scale human-labeled dataset for complex and cross-domain semantic parsing and text-to-{SQL} task}.
In: \beditor{\bsnm{Riloff}, \binits{E.}},
\beditor{\bsnm{Chiang}, \binits{D.}},
\beditor{\bsnm{Hockenmaier}, \binits{J.}},
\beditor{\bsnm{Tsujii}, \binits{J.}} (eds.)
\bbtitle{Proceedings of the 2018 Conference on Empirical Methods in Natural Language Processing},
pp. \bfpage{3911}--\blpage{3921}.
\bpublisher{Association for Computational Linguistics},
\blocation{Brussels, Belgium}
(\byear{2018}).
\doiurl{10.18653/v1/D18-1425} .
\burl{https://aclanthology.org/D18-1425}
\end{bchapter}
\endbibitem

%%% 26
\bibitem[\protect\citeauthoryear{Zhang et~al.}{2023}]{zhang2023predicting}
\begin{barticle}
\bauthor{\bsnm{Zhang}, \binits{Y.}},
\bauthor{\bsnm{An}, \binits{R.}},
\bauthor{\bsnm{Liu}, \binits{S.}},
\bauthor{\bsnm{Cui}, \binits{J.}},
\bauthor{\bsnm{Shang}, \binits{X.}}:
\batitle{Predicting and understanding student learning performance using multi-source sparse attention convolutional neural networks}.
\bjtitle{IEEE Transactions on Big Data}
\bvolume{9}(\bissue{01}),
\bfpage{118}--\blpage{132}
(\byear{2023})
\end{barticle}
\endbibitem

%%% 27
\bibitem[\protect\citeauthoryear{Zhang et~al.}{2024}]{zhang2024buildingcooperativeembodiedagents}
\begin{botherref}
\oauthor{\bsnm{Zhang}, \binits{H.}},
\oauthor{\bsnm{Du}, \binits{W.}},
\oauthor{\bsnm{Shan}, \binits{J.}},
\oauthor{\bsnm{Zhou}, \binits{Q.}},
\oauthor{\bsnm{Du}, \binits{Y.}},
\oauthor{\bsnm{Tenenbaum}, \binits{J.B.}},
\oauthor{\bsnm{Shu}, \binits{T.}},
\oauthor{\bsnm{Gan}, \binits{C.}}:
Building Cooperative Embodied Agents Modularly with Large Language Models
(2024).
\url{https://arxiv.org/abs/2307.02485}
\end{botherref}
\endbibitem

%%% 28
\bibitem[\protect\citeauthoryear{Zhang et~al.}{2020}]{zhang2020meta}
\begin{barticle}
\bauthor{\bsnm{Zhang}, \binits{Y.}},
\bauthor{\bsnm{Dai}, \binits{H.}},
\bauthor{\bsnm{Yun}, \binits{Y.}},
\bauthor{\bsnm{Liu}, \binits{S.}},
\bauthor{\bsnm{Lan}, \binits{A.}},
\bauthor{\bsnm{Shang}, \binits{X.}}:
\batitle{Meta-knowledge dictionary learning on 1-bit response data for student knowledge diagnosis}.
\bjtitle{Knowledge-Based Systems}
\bvolume{205},
\bfpage{106290}
(\byear{2020})
\end{barticle}
\endbibitem

%%% 29
\bibitem[\protect\citeauthoryear{Zhang et~al.}{2022}]{zhang2022multi}
\begin{botherref}
\oauthor{\bsnm{Zhang}, \binits{Y.}},
\oauthor{\bsnm{Liu}, \binits{S.}},
\oauthor{\bsnm{Qu}, \binits{X.}},
\oauthor{\bsnm{Shang}, \binits{X.}}:
Multi-instance discriminative contrastive learning for brain image representation.
Neural Computing and Applications,
1--14
(2022)
\end{botherref}
\endbibitem

%%% 30
\bibitem[\protect\citeauthoryear{Zhang et~al.}{2023}]{zhang2023federated}
\begin{botherref}
\oauthor{\bsnm{Zhang}, \binits{Y.}},
\oauthor{\bsnm{Wang}, \binits{Y.}},
\oauthor{\bsnm{Li}, \binits{Y.}},
\oauthor{\bsnm{Xu}, \binits{Y.}},
\oauthor{\bsnm{Wei}, \binits{S.}},
\oauthor{\bsnm{Liu}, \binits{S.}},
\oauthor{\bsnm{Shang}, \binits{X.}}:
Federated discriminative representation learning for image classification.
IEEE Transactions on Neural Networks and Learning Systems
(2023)
\end{botherref}
\endbibitem

%%% 31
\bibitem[\protect\citeauthoryear{Zhang et~al.}{2019}]{zhang2019editingbasedsqlquerygeneration}
\begin{botherref}
\oauthor{\bsnm{Zhang}, \binits{R.}},
\oauthor{\bsnm{Yu}, \binits{T.}},
\oauthor{\bsnm{Er}, \binits{H.Y.}},
\oauthor{\bsnm{Shim}, \binits{S.}},
\oauthor{\bsnm{Xue}, \binits{E.}},
\oauthor{\bsnm{Lin}, \binits{X.V.}},
\oauthor{\bsnm{Shi}, \binits{T.}},
\oauthor{\bsnm{Xiong}, \binits{C.}},
\oauthor{\bsnm{Socher}, \binits{R.}},
\oauthor{\bsnm{Radev}, \binits{D.}}:
Editing-Based SQL Query Generation for Cross-Domain Context-Dependent Questions
(2019).
\url{https://arxiv.org/abs/1909.00786}
\end{botherref}
\endbibitem

%%% 32
\bibitem[\protect\citeauthoryear{Zhong et~al.}{2020}]{zhong2020semanticevaluationtexttosqldistilled}
\begin{botherref}
\oauthor{\bsnm{Zhong}, \binits{R.}},
\oauthor{\bsnm{Yu}, \binits{T.}},
\oauthor{\bsnm{Klein}, \binits{D.}}:
Semantic Evaluation for Text-to-SQL with Distilled Test Suites
(2020).
\url{https://arxiv.org/abs/2010.02840}
\end{botherref}
\endbibitem

\end{thebibliography}
%% if required, the content of .bbl file can be included here once bbl is generated
%%\input sn-article.bbl

\end{document}